\documentclass[review]{elsarticle}

\usepackage{algorithmic}
\usepackage{array}
\usepackage{framed,multirow,booktabs}
\usepackage{graphicx,epsfig}
\usepackage{amssymb,amsmath,bm}
\usepackage{textcomp}
\usepackage{amsthm}
\usepackage{lineno,hyperref}
\usepackage{xcolor}
\usepackage{cleveref}
\usepackage[section]{placeins}
\usepackage{lineno,hyperref}

\modulolinenumbers[5]
\journal{Journal of \LaTeX\ Templates}

\begin{document}

\begin{frontmatter}

\title{Split, Embed and Merge: An accurate table structure recognizer}

\author{Zhenrong Zhang, Jianshu Zhang, Jun Du and Fengren Wang}

\begin{abstract}
	Table structure recognition is an essential part for making machines understand tables. Its main task is to recognize the internal structure of a table. However, due to the complexity and diversity in their structure and style, it is very difficult to parse the tabular data into the structured format which machines can understand, especially for complex tables. In this paper, we introduce Split, Embed and Merge (SEM), an accurate table structure recognizer. SEM is mainly composed of three parts, splitter, embedder and merger. In the first stage, we apply the splitter to predict the potential regions of the table row/column separators, and obtain the fine grid structure of the table. In the second stage, by taking a full consideration of the textual information in the table, we fuse the output features for each table grid from both vision and text modalities. Moreover, we achieve a higher precision in our experiments through providing additional textual features. Finally, we process the merging of these basic table grids in a self-regression manner. The corresponding merging results are learned through the attention mechanism. In our experiments, SEM achieves an average F1-Measure of $97.11\%$ on the SciTSR dataset which outperforms other methods by a large margin. We also won the first place of complex tables and third place of all tables in Task-B of ICDAR 2021 Competition on Scientific Literature Parsing. Extensive experiments on other publicly available datasets further demonstrate the effectiveness of our proposed approach.
\end{abstract}


\end{frontmatter}


\section{Introduction}
	In this age of knowledge and information, documents are a very important source of information for many different cognitive processes such as knowledge database creation, optical character recognition (OCR), graphic understanding, document retrieval, etc. Automatically processing the information embedded in these documents is crucial. Numerous efforts have been made in the past to automatically extract the relevant information from documents~\cite{DSSE, DeCNT, DeepDeSRT}. As a particular entity, the tabular structure is very commonly encountered in documents. These tabular structures convey some of the most important information in a very concise form. Therefore, they are extremely prevalent in domains like finance, administration, research, and even archival documents. Table structure recognition (TSR) aims to recognize the table internal structure to the machine readable data mainly presented in two formats: logical structure and physical structure~\cite{SurveyTable}. More concretely, logical structure only contains every cell’s row and column spanning information, while the physical one additionally contains bounding box coordinates of cells. As a result, table structure recognition as a precursor to contextual table understanding will be useful in a wide range of applications~\cite{DSSE, DeCNT, DeepDeSRT}.

\begin{figure*}[htb]
	\centerline{\includegraphics[width=1.\linewidth]{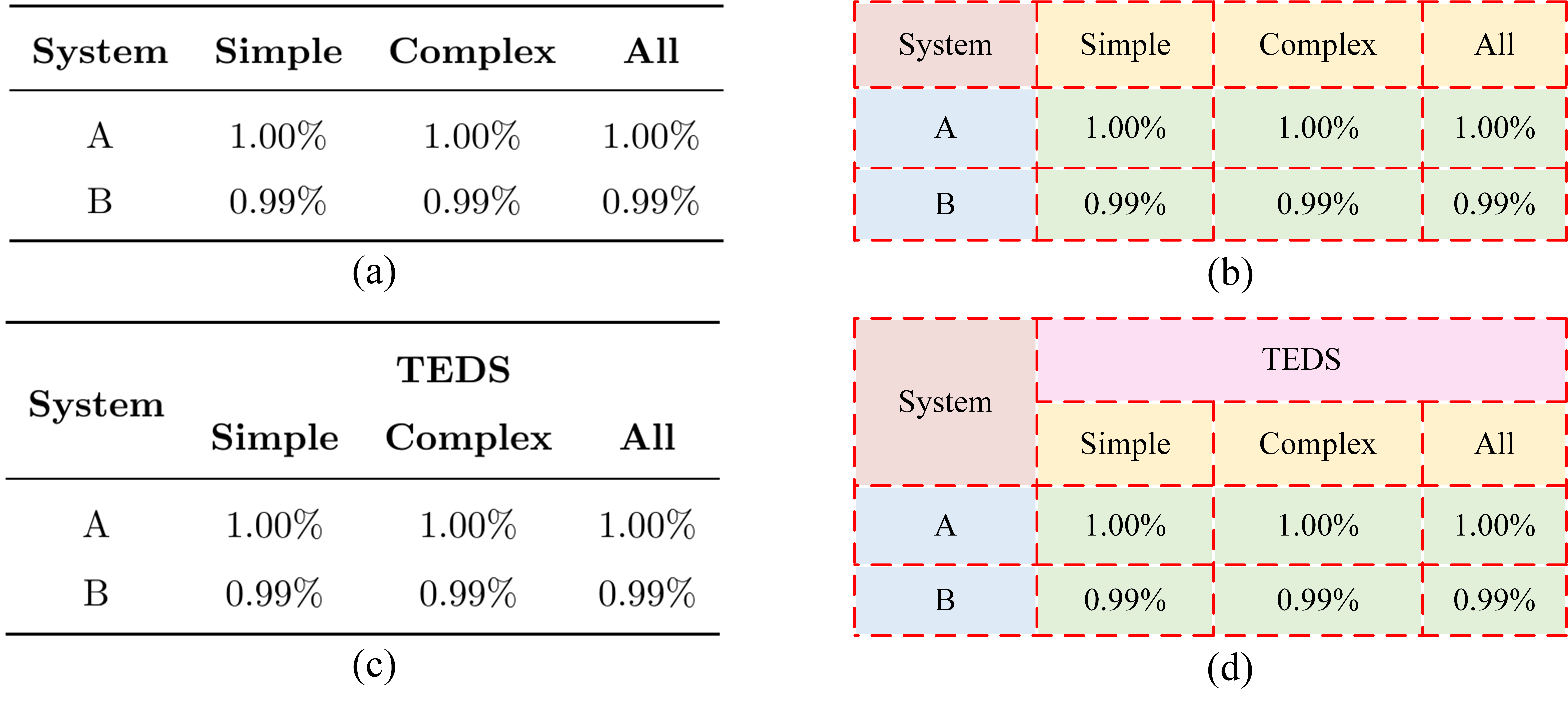}}
	\caption{An intuitive comparison between simple and complex tables. The example of the simple table is shown in (a), and (b) is its real structure. The example of the complex table is shown in (c), and (d) is its real structure. Note that in (d), the cells with the contents of “System” and “TEDS” occupy multiple rows or multiple columns, so it is a complex table.}
	\label{simple vs. complex table example}
\end{figure*}

Table structure recognition is a challenging problem due to the complex structure and high variability in table layouts. A spanning cell is a table cell that occupies at least two rows or columns. If a table contains spanning cells, it is called a complex table, as shown in Figure~\ref{simple vs. complex table example}. Although significant efforts have been made in the past to recognize the internal structure of tables through an automated process~\cite{DeepDeSRT, EDD, TabStructNet}, most of these methods~\cite{DeepDeSRT, SemanticSegmTSR} only focus on simple tables and are hard to accurately recognize the structure of complex tables. The spanning cells usually contain more important semantic information than other simple cells, because they are more likely to be table headers in a table. The table header is crucial to understand the table. Therefore, more needs to be done for recognizing the structure of complex tables. 

Recently, many works~\cite{VQA-M4C, ImageCaption-DVSA, DSSE} have demonstrated the significant impact of using visual and textual representations in a joint framework. However, most previous methods~\cite{DeepDeSRT, TabStructNet, GraphTSR} in table structure recognition only use the spatial or visual features without considering the textual information of each table cell. The structures of some tables have a certain ambiguity from the visual appearance, especially for table cells which contain multi-line contents. Therefore, to recognize the table structure accurately, it is inevitable to take advantage of the cross-modality nature of visually-rich table images, where visual and textual information should be jointly modeled. In our work, we design vision module and text module in our embedder to extract visual features and textual features, respectively, and achieve a higher recognition accuracy.

Most existing literature~\cite{GraphTSR, ReS2TIM, DGCNN} on table structure recognition depends on extraction of meta-information from the pdf document or the OCR models to extract low-level layout features from the image. Nevertheless, these methods fail to extend to scanned documents due to the absence of meta-information or errors made by the OCR, when there is a wide variety in table layouts and text organization. In our work, we address the problem of table structure recognition by directly operating over table images with no dependency on meta-information or OCR.

In this study, we introduce Split, Embed and Merge (SEM), an accurate table structure recognizer as shown in Figure~\ref{overview}. Considering that the table is composed of a set of table cells and each table cell is composed of one or more basic table grids, we deem table grids as the basic processing units in our framework. Therefore, we design the pipeline of SEM as follows: 1) divide table into basic table grids 2) merge them to recover the table cells. The final table structure can be obtained by parsing all table cells. As a consequence, SEM mainly has three components: splitter, embedder and merger. The splitter, which is actually a fully convolutional network (FCN)~\cite{FCN}, is first applied to predict the fine grid structure of the table as shown in the upper-right of Figure~\ref{overview}. The embedder as a feature extractor embeds vision and plain text contained in a table grid into a feature vector. More specifically, we use the RoIAlign~\cite{MaskRCNN} to extract the visual features from the output of the backbone, and extract textual features using the off-the-shelf recognizer~\cite{WAP} and the pretrained BERT~\cite{Bert} model. Finally, the merger which is a Gated Recurrent Unit (GRU) decoder will predict the gird merged results step by step based on the grid-level features extracted by the embedder. For each predicted merged result, the attention mechanism built into the merger scans the entire grid elements and predicts which grids should be merged at the current step. The proposed method can not only process simple tables well, but also complex tables. The ambiguity problem of the table structure recognition based on visual appearance can be alleviated through our embedder. Moreover, SEM is able to directly operate over table images, which enhances the applicability of the system (to both PDFs and images).

The main contributions of this paper are as follows:
\begin{itemize}
	
	\item 
	We present an accurate table structure recognizer, Split, Embed and Merge (SEM), to recognize the table structure. The designed merger can accurately predict table structure based on the fine grid structure of the table. This proposed new method can not only process simple tables well, but also complex tables.
	
	\item
	We demonstrate that jointly modeling the visual and textual information in the table will further boost model performance. Through visualization in experiments, the ambiguity problem of the table structure recognition can be alleviated based on our multimodality features.
	
	\item 
	Based on our proposed method, we won the first place of complex tables and the third place of all tables in Task-B of ICDAR 2021 Competition on scientific literature parsing. In addition, we also achieved the results with an average F1-Measure of $97.11\%$ and $95.72\%$ in SciTSR and SciTSR-COMP datasets, respectively, demonstrating the effectiveness of our method.
	
\end{itemize}

\section{Related Work}
\subsection{Table Structure Recognition}
Analyzing tabular data in unstructured documents mainly focuses on three problems: i) table detection: localizing the bounding boxes of tables in documents~\cite{Icdar-2013-Comp, Icdar-2019-Comp}, ii) table structure recognition: parsing only the structural (row and column layout) information of tables~\cite{DeepDeSRT, TabStructNet, SPLERGE}, and iii) table recognition: parsing
both the structural information and content of table cells~\cite{EDD}. In this study, we mainly focus on table structure recognition. Most early proposed methods~\cite{TSR-h1, TSR-h2, TSR-h3} are based on heuristics. While these methods were primarily dependent on hand-crafted features and heuristics (horizontal and vertical ruling lines, spacing and geometric analysis). 

Due to the rapid development of deep learning and the massive amounts of tabular data in documents on the Web, many deep learning-based methods~\cite{DeepDeSRT, EDD, TabStructNet, GraphTSR, SemanticSegmTSR}, which are robust to the input type (whether being scanned images or native digital), have also been presented to understand table structures. These also do not make any assumptions about the layouts, are data-driven, and are easy to fine-tune across different domains. \cite{DeepDeSRT, SemanticSegmTSR} utilize recently published insights from semantic segmentation~\cite{FCN} research for identifying rows, columns, and cell positions within tables to recognize table structures. However, \cite{DeepDeSRT, SemanticSegmTSR} do not consider the complex tables containing spanning cells, so that they cannot handle the structure recognition of complex tables well. GraphTSR~\cite{GraphTSR} proposes a novel graph neural network for recognizing the table structure in PDF files and can recognize the structure of complex tables. GraphTSR takes the table cells as input which means that it fails to generalize well due to the absence of meta-information or errors made by the OCR. EDD~\cite{EDD} treats table structure recognition as a task similar to img2latex~\cite{WAP, TreeDecoder}. EDD directly generates the HTML tags that define the structure of the table through an attention-based structure decoder. \cite{TabStructNet} presents the TabStructNet for table structure recognition that combines cell detection and interaction modules to localize the cells and predict their row and column associations with other detected cells. Compared with the aforementioned methods, our method SEM not only takes table images as input, but also can recognize the structure of complex tables well.

\subsection{Attention Mechanisms}
Given a query element and a set of key elements, an attention function can adaptively aggregate the key contents according to attention weights, which measure the compatibility of query-key pairs. The attention mechanisms as an integral part of models enable neural networks to focus more on relevant elements of the input than on irrelevant parts. They were first studied in natural language processing (NLP), where encoder-decoder attention modules were developed to facilitate neural machine translation~\cite{MachineTranslation-1, MachineTranslation-2, ConvolutionalSequence, Transformer}. In
particular, self-attention, also called intra-attention, is an attention mechanism relating different positions of a single sequence in order to compute a representation of the sequence. Self-attention has been used successfully in a variety of tasks including reading comprehension, abstractive summarization, and textual entailment. The landmark work, Transformer~\cite{Transformer}, presents the transduction model relying entirely on self-attention to compute representations of its input and output, and substantially surpasses the performance of past work. 

The success of attention modeling in NLP~\cite{MachineTranslation-1, ConvolutionalSequence, Transformer} has also led to its adoption in computer vision such as object detection~\cite{DETR, AttentionForDetection}, semantic segmentation~\cite{AttentionForSegmentation-1, AttentionForSegmentation-2}, image captioning~\cite{AttentionForImageCaption} and text recognition~\cite{WAP, MaskTextSpotterV2}, etc. DETR~\cite{DETR} completes the object detection by adoptting an encoder-decoder architecture based on transformers~\cite{Transformer} to directly predict a set of object bounding boxes. In order to capture contextual information, especially in the long range, \cite{AttentionForSegmentation-2} proposes the point-wise spatial attention network (PSANet) to aggregate long-range contextual information in a flexible and adaptive manner. Mask TextSpotter v2~\cite{MaskTextSpotterV2} applies a spatial attentional module for text recognition, which alleviates the problem of character-level annotations and improves the performance significantly. In our work, we apply the transformers to capture the long-range dependencies on grid-level featuers and build attention mechanisms into our merger to predict which gird elements should be merged together to recover table cells.

\subsection{Multimodality}
Several joint learning tasks such as image captioning~\cite{DSEN, ImageCaption-DVSA}, visual question answering~\cite{VQA-1, VQA-2, VQA-M4C}, and document semantic structure extraction~\cite{DSSE} have demonstrated the significant impact of using visual and textual representations in a joint framework. \cite{ImageCaption-DVSA} aligned parts of visual and language modalities through a common, multimodal embedding, and used the inferred alignments to learn to generate novel descriptions of image regions. \cite{VQA-M4C} proposed a novel model, Multimodal Multi-Copy Mesh (M4C), for the TextVQA task based on a multimodal transformer architecture accompanied by a rich representation for text in images and achieved the state-of-the-art. \cite{DSSE} considered document semantic structure extraction as a pixel-wise segmentation task, and presented a unified model, Multimodal Fully Convolutional Network (MFCN). MFCN classifies pixels based not only on their visual appearance, as in the traditional page segmentation task, but also on the content of underlying text. In our work, we take a full consideration of the plain text contained in table images, and design the embedder to extract both visual and textual features at the same time. The experiments also prove that more accurate results will be obtained when providing additional textual information on visual clues.

\begin{figure*}[htb]
	\centerline{\includegraphics[width=1.\linewidth]{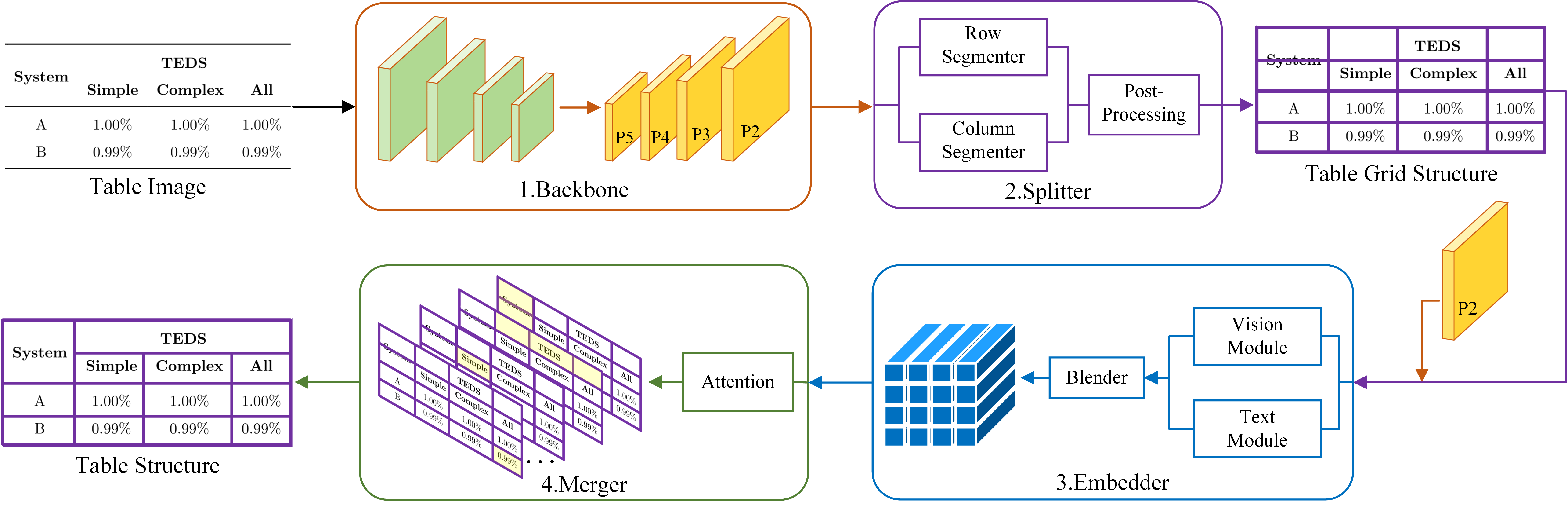}}
	\caption{\textbf{SEM pipeline} The backbone is applied to extract the feature maps from the table image. The splitter uses the backbone features to predict a set of basic table grids. The embedder extracts the region features corresponding to each basic table gird. The merger predicts which grid elements need to be merged to recover the table cells.}
	\label{overview}
\end{figure*}

\section{Method}
The overall pipeline of our system is shown in Figure~\ref{overview}. The modified ResNet-34~\cite{ResNet} with FPN~\cite{FPN} as our backbone is first applied to the input table image to extract multi-level feature maps. The splitter takes the output of the backbone as input and predicts the fine grid structure of the table. The table grid structure is in the form of row and column separators that span the entire image as shown in the upper-right of Figure~\ref{overview}. The following embedder extracts the feature representation of each basic table grid. Finally, based on the grid-level features extracted by the embedder, the merger with the attention mechanism will predict which grids should be merged step by step. The table structure can be recovered based on the merged results. In the following subsections, three main components in our system, namely, the splitter, the embedder and the merger, will be elaborated.

\begin{figure*}[htb]
	\centerline{\includegraphics[width=1.\linewidth]{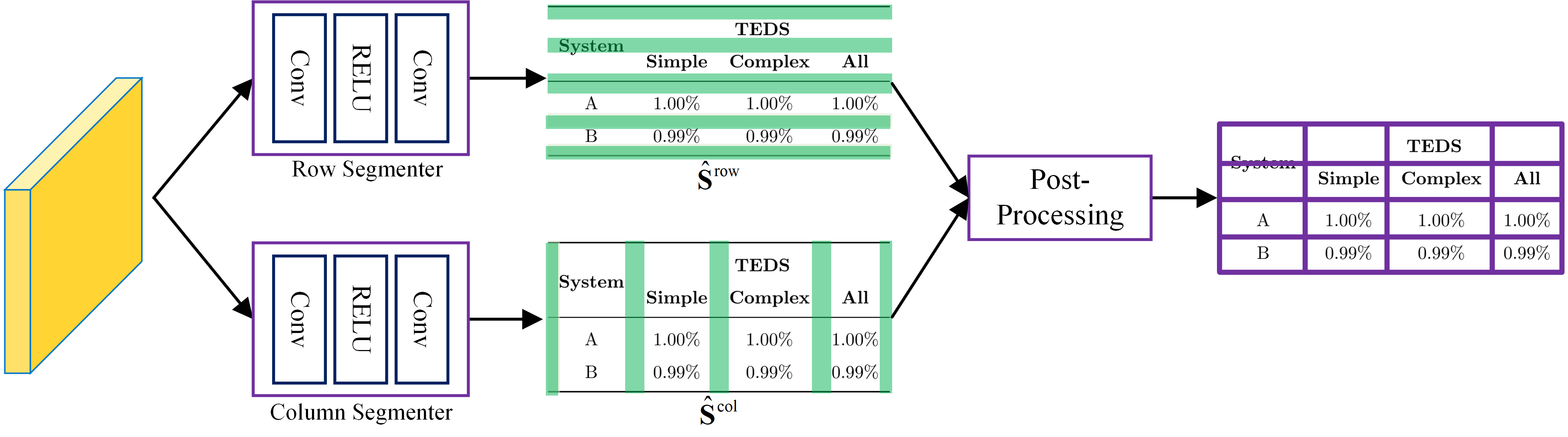}}
	\caption{The illustration of the splitter. The splitter takes a feature map as input and predicts the potential regions of the table row/column separators, which are the green masks in the table images. The following post-processing is used to extract the basic table grids according to the segmentation results from the segmenters.}
	\label{splitter}
\end{figure*}

\subsection{Splitter}
Different from the method in~\cite{EDD}, performing table structure prediction on the image-level features, we believe that using table grids as the basic processing units will be more reasonable, so we design the splitter to predict the basic table grid pattern. Inspired by the segmentation-based methods~\cite{TextMountain, MaskTextSpotter} in the field of text detection and the FCN~\cite{FCN} in image segmentation, we refer to the potential regions of the table row/column separators as the foreground and design the splitter which contains two separate row/column segmenters to predict the table row/column separator maps $\hat{\mathbf{S}}^{\text{row}}$/$\hat{\mathbf{S}}^{\text{col}}$ as shown in Figure~\ref{splitter}. ${{\hat{\mathbf{S}}}^{\text{row}}}/\hat{\mathbf{S}}^{\text{col}} \in {\mathbb{R}^{H \times W}}$ and ${H \times W}$ is the size of the input image. 

Each segmenter is actually the fully convolutional network which contains a convolutional layer, ReLU and a convolutional layer. As some table row/column separator regions are quite slender, it is important to ensure segmentation results have a high resolution. The kernel size and the stride of each convolutional layer in the segmenter is set to $3 \times 3$ and $1$, respectively, which keeps the same spatial resolution of the input and the output. Moreover, we modify the ResNet-34 by setting the stride of the first convolutional layer with $7 \times 7$ kernel size to $1$, and remove the adjacent pooling layer to guarantee the resolution of the lowest-level feature map is consistent with the input image. We strongly believe that rich semantics extracted by deeper layers can help with obtaining more accurate segmentation results, so we add a top-down path~\cite{FPN} in our backbone to enrich semantics in feature maps. Finally, the backbone generates a feature pyramid with four feature maps \{$\text{P2}$, $\text{P3}$, $\text{P4}$, $\text{P5}$\}, whose final output strides are $1$, $2$, $4$, $8$, respectively. The number of channels in the feature maps is $D$. We take $\text{P2}$ as the input of the splitter.

The loss function is defined as follows:
\begin{align}
	& \mathcal{L}_s^{\text{row}} = \sum\limits_{j = 1}^H {\sum\limits_{i = 1}^W {\frac{{L(\hat{\text{S}}_{i,j}^{\text{row}}, \text{S}_{i,j}^{\text{row}})}}{{\sum\limits_{j = 1}^H {\sum\limits_{i = 1}^W {\text{S}_{i,j}^{\text{row}}} } }}} }
	\label{seg-row-loss} \\
	& \mathcal{L}_s^{\text{col}} = \sum\limits_{j = 1}^H {\sum\limits_{i = 1}^W {\frac{{L(\hat{\text{S}}_{i,j}^{\text{col}}, \text{S}_{i,j}^{\text{col}})}}{{\sum\limits_{j = 1}^H {\sum\limits_{i = 1}^W {\text{S}_{i,j}^{\text{col}}} } }}} }
	\label{seg-col-loss}
\end{align}
in which
\begin{align}
	\hspace{-1.cm}
	{L(x,y)} = -(y\log (\sigma(x)) + (1-y)\log (1 - \sigma(x)))
	\label{BCE}
\end{align}
where $\mathbf{S}^{\text{row}}$/$\mathbf{S}^{\text{col}}$ denotes the ground-truth of the table row/column separator map. $\text{S}_{i,j}^{\text{row}}$/$\text{S}_{i,j}^{\text{col}}$ is $1$ if the pixel in $i^{th}$ column and $j^{th}$ row belongs to the table row/column separator region, otherwise 0. The $\sigma$ is the sigmoid function.

The goal of our post-processing is to extract table row/column lines from the table row/column separator map as shown in Figure~\ref{splitter}. Specifically, we first apply the sigmoid function to the predicted segmentation map $\hat{\mathbf{S}}^{\text{row}}$/$\hat{\mathbf{S}}^{\text{col}}$ and average them by column/row size to obtain the $\bar{\mathbf{S}}^{\text{row}}$/$\bar{\mathbf{S}}^{\text{col}}$ as illustrated in Eq.~\ref{Mean-row}~\ref{Mean-col}, where ${{\bar{\mathbf{S}}}^{\text{row}}} \in {\mathbb{R}^{H \times 1}}$ and ${{\bar{\mathbf{S}}}^{\text{col}}} \in {\mathbb{R}^{1 \times W}}$. Then we binarize the $\bar{\mathbf{S}}^{\text{row}}$/$\bar{\mathbf{S}}^{\text{col}}$ into $\tilde{\mathbf{S}}^{\text{row}}$/$\tilde{\mathbf{S}}^{\text{col}}$, $\tilde{\text{S}}_j^{\text{row}}$/$\tilde{\text{S}}_i^{\text{col}}$ = 1 indicating this row/column is a potential table line. For the block that is equal to 1 in $\tilde{\mathbf{S}}^{\text{row}}$/$\tilde{\mathbf{S}}^{\text{col}}$, we select the row/column with the maximum value of the corresponding block in $\bar{\mathbf{S}}^{\text{row}}$/$\bar{\mathbf{S}}^{\text{col}}$ as the final table row/column line.
\begin{align}
	& \bar{\text{S}}_j^{\text{row}}=\frac{1}{W}\sum_i^W{\sigma \left({\hat{\text{S}}}_{i,j}^{\text{row}} \right)} 
	\label{Mean-row}\\
	& \bar{\text{S}}_i^{\text{col}}=\frac{1}{H}\sum_j^H{\sigma \left({\hat{\text{S}}}_{i,j}^{\text{col}} \right)}
	\label{Mean-col}
\end{align}

We can easily obtain a set of bounding boxes $\mathbf{G}$ of table grids from the table row/column lines. $\mathbf{G} \in \mathbb{R}^{(M\times N)\times 4}$, where $M$, $N$ are the number of rows and columns occupied by the table grid structure, respectively. More specifically, each bounding box can be precisely defined by $(x_1, y_1, x_2, y_2)$, where $(x_1, y_1)$ corresponds to the position of the upper left in the bounding box, and $(x_2, y_2)$ represents the position of the lower right. 

\subsection{Embedder}

\begin{figure*}[htb]
	\centerline{\includegraphics[width=1.\linewidth]{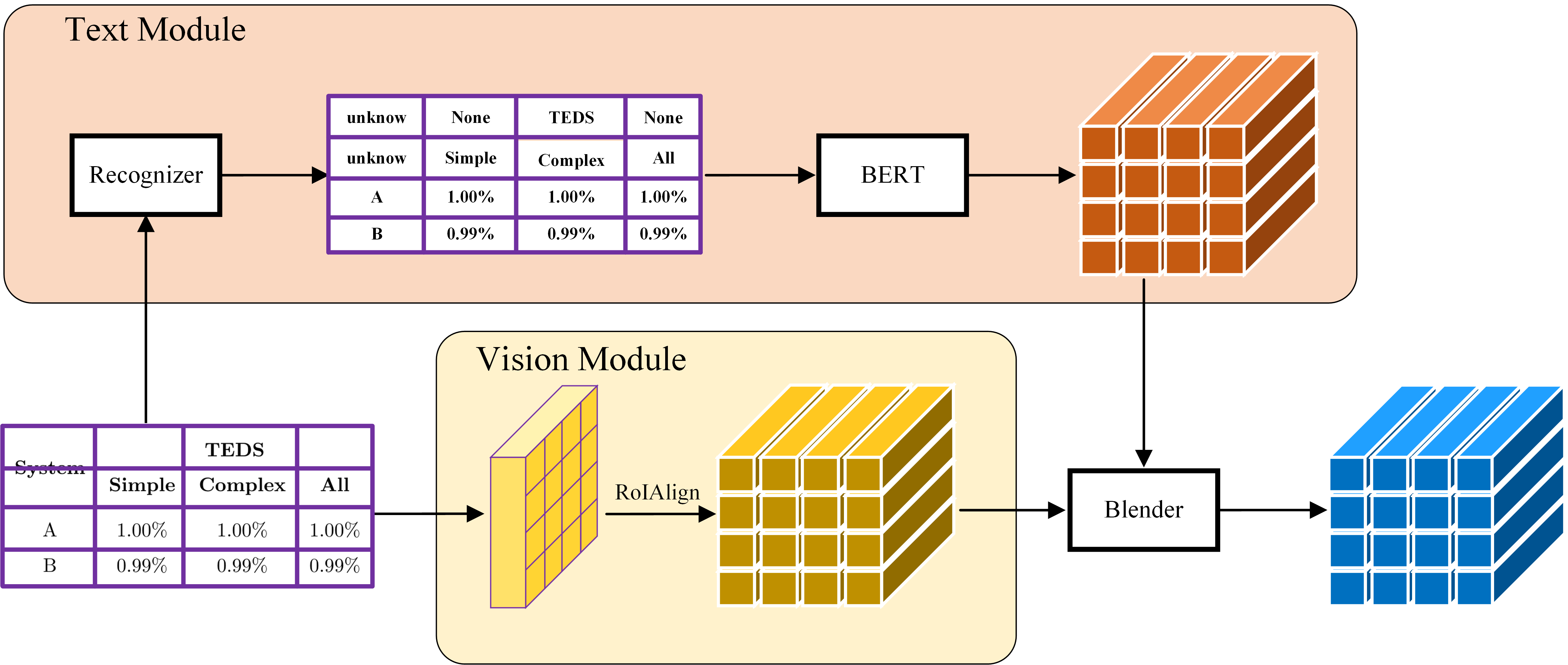}}
	\caption{The illustration of the embedder. It is composed of vision module (VM), text module (TM) and blender module (BM). The embedder extracts the gird-level visual and textual features from VM and TM, respectively. Finally, the BM fuses the both features.}
	\label{embedder}
\end{figure*}

The embedder aims to extract the feature representations of each grid. \cite{ImageCaption-DVSA, VQA-M4C} have demonstrated the effectiveness of taking advantage of the multi-modality. Different from the previous table structure recognition methods~\cite{EDD,TabStructNet,SPLERGE} which mostly recover the table structure based on the visual modality, we fuse the output features for each basic table grid from both visual and textual modalities. Therefore, we design the vision module and text module in the embedder to extract visual features $\mathbf{E}^v$ and textual features $\mathbf{E}^t$, respectively, and fuse both features to produce the final grid-level features $\mathbf{E}$ through the blender module. $\mathbf{E}^v\in \mathbb{R}^{(M\times N)\times D}$, $\mathbf{E}^t\in \mathbb{R}^{(M\times N)\times D}$ and $\mathbf{E}\in \mathbb{R}^{(M\times N)\times D}$, where $D$ represents the number of feature channels. 

As shown in Figure~\ref{embedder}, the vision module takes the image-level feature map $\text{P2}$ from the FPN and the well-divided table grids $\mathbf{G}$ obtained from the splitter as input. It applies the RoIAlign~\cite{MaskRCNN} to pool a fixed size $R\times R$ feature map $\mathbf{\hat E}_{i}^{v}$ for each table grid.
\begin{equation}
	\mathbf{\hat E}_i^v = \mbox{RoIAlign}_{R \times R}(\text{P2},{\mathbf{G}_i}) \quad \forall i = \{ 1,...,M\times N\} 
\end{equation}
where $\mathbf{\hat E}_i^v \in \mathbb{R}^{R\times R\times D}$. The final visual features $\mathbf{E}_{i}^{v}$ are obtained according to:
\begin{equation}
	\mathbf{E}_i^v = \text{FFN}(\mathbf{\hat E}_i^v) \quad \forall i = \{ 1,...,M\times N\} 
\end{equation}
in which
\begin{equation}
	{\text{FFN}(\mathbf{x})} = \max (0, \mathbf{x}\mathbf{W}_1 + \mathbf{b}_1)\mathbf{W}_2 + \mathbf{b}_2
	\label{FFN}
\end{equation}
where \text{FFN}~\cite{Transformer} is actually two linear transformations with a ReLU activation in between. $\mathbf{x}\in \mathbb{R}^{d_{\text{in}}}$, $\mathbf{W}_1 \in \mathbb{R}^{d_{\text{in}} \times d_{\text{ff}}}$, $\mathbf{b}_1 \in \mathbb{R}^{d_{\text{ff}}}$, $\mathbf{W}_2 \in \mathbb{R}^{d_{\text{ff}} \times d_{\text{out}}}$, $\mathbf{b}_2 \in \mathbb{R}^{d_{\text{out}}}$. The dimensionality of input and output is $d_{\text{in}}$ and $d_{\text{out}}$, and the inner-layer has dimensionality $d_{\text{ff}}$. Here we set $d_{\text{ff}} = d_{\text{out}}$ in default.

The table image is both visually-rich and textual-rich, so it is necessary to make full use of the textual information in the table to achieve a more accurate table structure recognizer. As shown in the text module of Figure~\ref{embedder}, we apply the off-the-shelf recognizer~\cite{WAP} to obtain a sequence of $M \times N$ contents for all table grids, and embed contents into corresponding feature vectors $\mathbf{\hat E}^{t}$ using a pretrained BERT model~\cite{Bert}. $\mathbf{\hat E}^t\in \mathbb{R}^{(M\times N)\times B}$, where $B$ is the feature vector dimension of the BERT. It's worth noting that both the recognizer and the BERT do not update the parameters during the training phase. The final textual features $\mathbf{E}^t$ are obtained by applying \text{FFN} again to fine-tune the extracted textual features $\mathbf{\hat E}^{t}$ to make it more suitable for our network.
\begin{equation}
	\mathbf{E}_i^t = \text{FFN}(\mathbf{\hat E}_i^t) \quad \forall i = \{ 1,...,M\times N\} 
\end{equation}

The blender module in Figure~\ref{embedder} is to fuse the visual features $\mathbf{E}^{v}$ and textual features $\mathbf{E}^{t}$, and its specific process is as follows:

1) For each basic table grid, we first obtain the intermediate results ${{{\mathbf{\hat E}}}_i}$ according to :
\begin{equation}
	{{{\mathbf{\hat E}}}_i} = \text{FFN}(\left[{\begin{array}{*{20}{c}} {{\mathbf{E}}_i^v}\\
			{{\mathbf{E}}_i^t} \end{array}} \right]) \quad \forall i \in \left[ {1,...,M\times N} \right]
\end{equation}
where $\left[  \cdot  \right]$ is the concatenation operation. The input and output dimensionality of the \text{FFN} is $2D$ and $D$, respectively.

2) So far, the features of each basic table grid are still independent of each other, especially for textual features. Therefore, we introduce the transformer~\cite{Transformer} to capture long-range dependencies on table grid elements. We take the features ${{{\mathbf{\hat E}}}}$ as query, key and value which are required by the transformer. The output of the transformer as final grid-level features $\mathbf{E}$ have a global receptive field.

\begin{figure*}[htb]
	\centerline{\includegraphics[width=1.\linewidth]{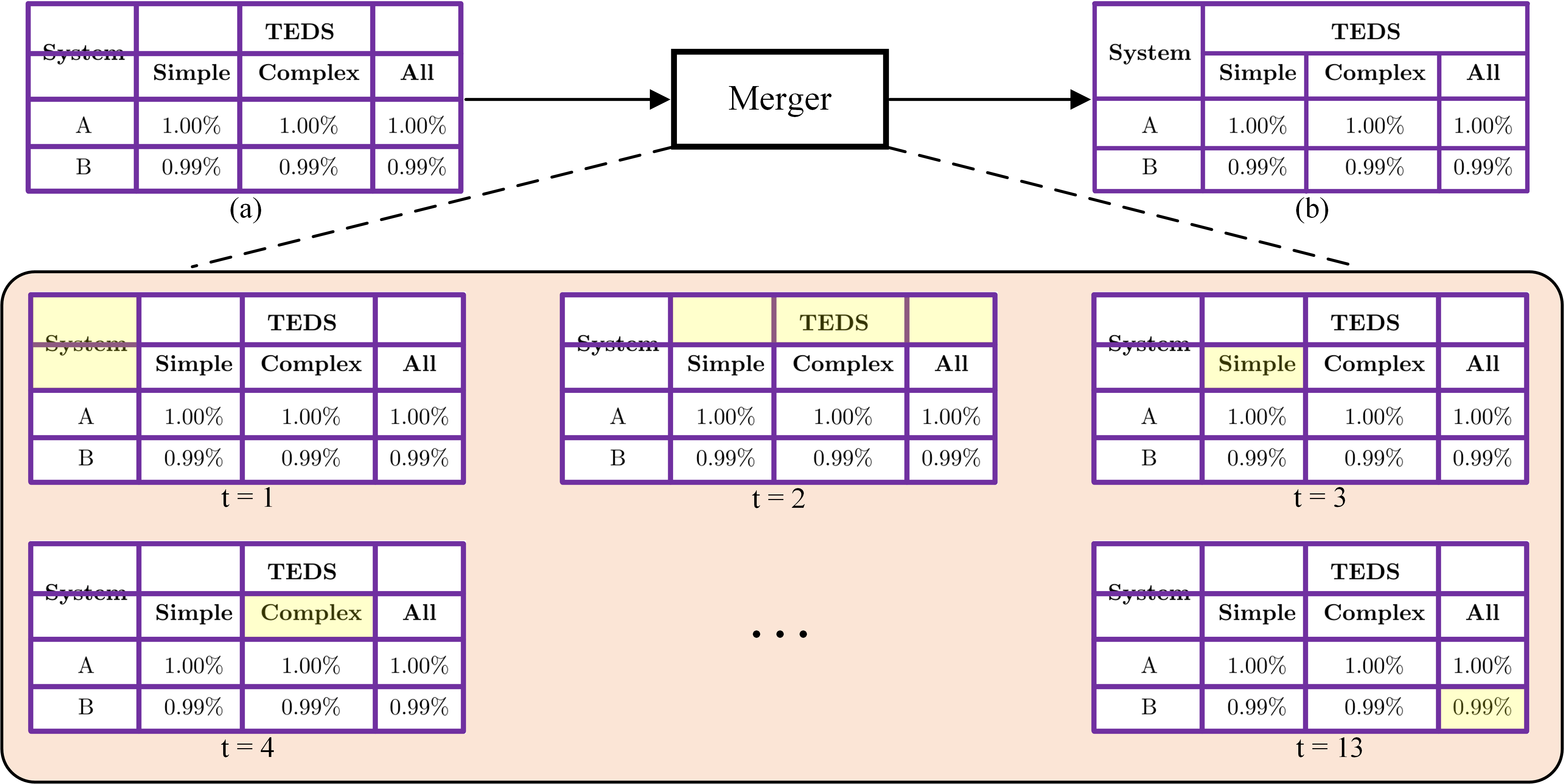}}
	\caption{The illustration of the merger. The yellow masks in lower part indicate which table grid elements should be merged in each time step.}
	\label{mergerlabel}
\end{figure*}

\subsection{Merger}

The merger is an RNN that takes the grid-level features $\mathbf{E}$ as input and produces a sequence of merged maps $\mathbf{M}$ as shown in Figure~\ref{mergerlabel}.
Here we choose the Gated Recurrent Unit (GRU)~\cite{GRU}, an improved version of simple RNN. 
\begin{equation}
	\mathbf{M} = \{\mathbf{m}_1, \mathbf{m}_2,...,\mathbf{m}_C\}
\end{equation}
where $C$ is the length of a predicted sequence. Each merged map $\mathbf{m}_t$ is a $(M\times N)$-dimension vector, the same size as the element of $\mathbf{E}$, and the value of each grid element $m_{ti}$ is $1$ or $0$, indicating whether the $i^{th}$ grid element belongs to the $t^{th}$ cell or not. The cells that span multiple rows or columns can be recovered according to $\mathbf{M}$.

\begin{figure*}[htb]
	\centerline{\includegraphics[width=1.\linewidth]{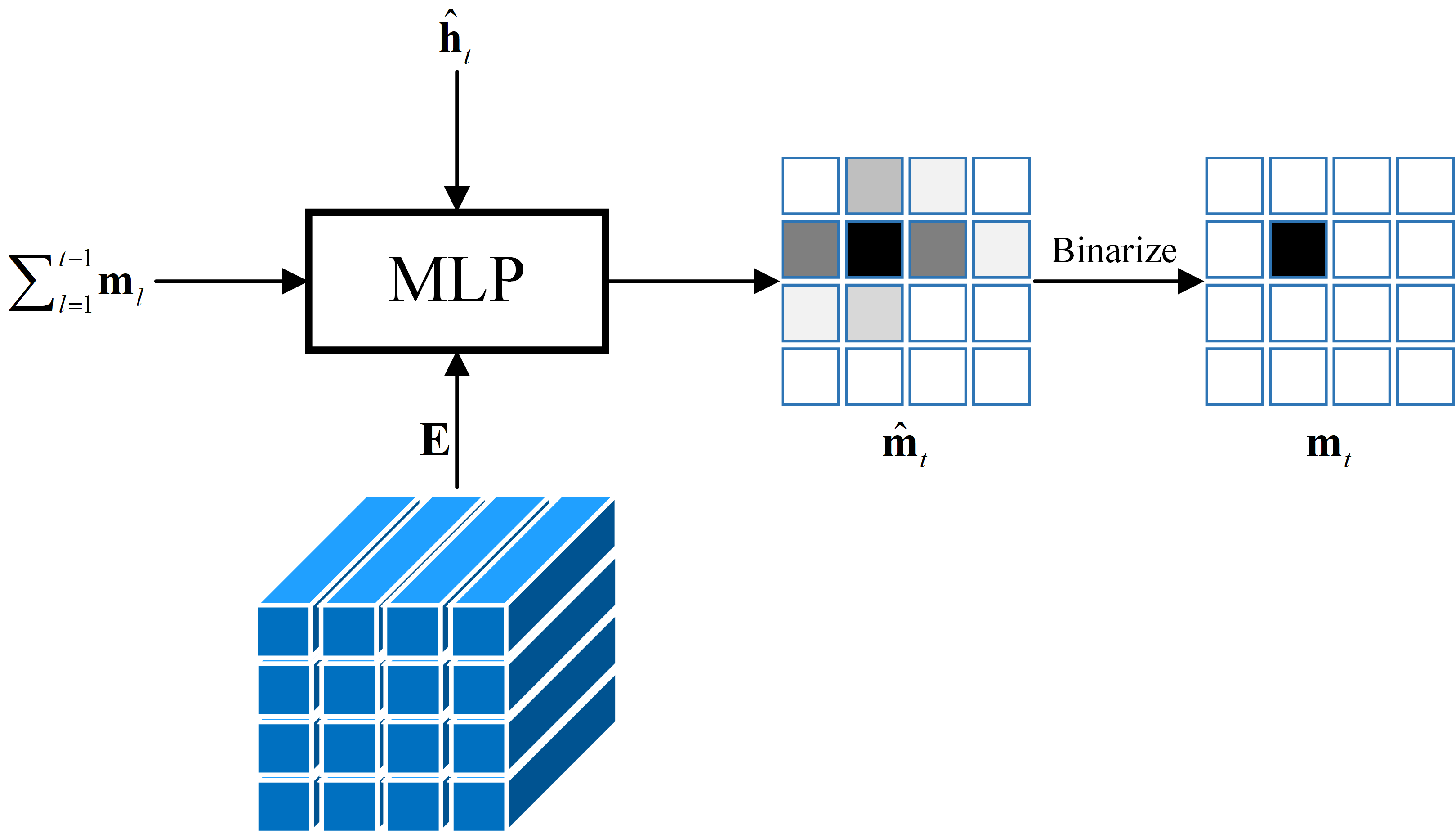}}
	\caption{The illustration of the attention mechanism. The prediction of current hidden state $\mathbf{\hat{h}}_t$ and the grid-level features $\mathbf{E}$ is used as query and key, respectively.}
	\label{fatt}
\end{figure*}

Inspired by the successful applications of attention mechanism in img2latex~\cite{WAP,TAP}, text recognition~\cite{ASTER,SEED}, machine translation~\cite{Transformer}, etc., we build the attention mechanism into our merger and achieve promising results. For the merged map $\mathbf{m}_t$ decoding, we compute the prediction of current hidden state $\mathbf{\hat{h}}_t$ from previous context vector $\mathbf{c}_{t-1}$ and its hidden state $\mathbf{h}_{t-1}$:
\begin{equation}
	\mathbf{\hat{h}}_t = \textrm{GRU}(\mathbf{c}_{t - 1}, \mathbf{h}_{t - 1})
\end{equation} 
Then we employ an attention mechanism with $\mathbf{\hat{h}}_t$ as the query and grid-level features $\mathbf{E}$ as both key and the value:
\begin{align}
	& {{\mathbf{m}}_t} = {f_{att}}({\mathbf{E}},{\mathbf{\hat{h}}_t})\\
	& {{\mathbf{c}}_t} = \frac{{{{\mathbf{m}}_t}}}{{{{\left\| {{{\mathbf{m}}_t}} \right\|}_1}}}{\mathbf{E}}
\end{align}
where $||\cdot||_1$ is the vector 1-norm. As shown in Figure~\ref{fatt}, we design $f_{att}$ function as follows:
\begin{align}
	& {\mathbf{F}} = {\mathbf{Q}} * \sum\nolimits_{l = 1}^{t - 1} {{{\mathbf{m}}_l}} \\
	& {{\hat m}_{ti}} = \bm{\nu}^{\rm T} \tanh ({{\mathbf{W}}_{att}}{\mathbf{\hat{h}}_t} + {{\mathbf{U}}_{att}}{{\mathbf{e}}_i} + {{\mathbf{U}}_F}{{\mathbf{f}}_i})\\
	& {m_{ti}} = \text{Binarize}({{\hat m}_{ti}})
\end{align}
in which
\begin{align}
	\hspace{-1.cm}
	{\text{Binarize}(x)} =
	\begin{cases}
		1 & \quad \text{if } \sigma (x) > 0.5 \\
		0 & \quad \text{otherwise}
	\end{cases}
	\label{Binarize}
\end{align}
where $*$ denotes a convolution layer, $\sum\nolimits_{l = 1}^{t - 1} {{{\mathbf{m}}_l}}$ denotes the sum of past determined grids, ${\hat m}_{ti}$ denotes the output energy, ${{\mathbf{f}}_i}$ denotes the element of $\mathbf{F}$, which is used to help append the history information into standard attention mechanism. It's worth noting that the attention mechanism is completed on the grid-level features. For each cell, it is quite clear which grid elements belong to it. Therefore, unlike the previous methods~\cite{WAP,TreeDecoder} using the softmax to obtain the attention probability, we use the $\text{Binarize}$ Eq.~\ref{Binarize} to calculate. Moreover, we find that the model is difficult to converge when using the softmax.

With the context vector $\mathbf{c}_{t}$, we compute the current hidden state:
\begin{equation}
	\mathbf{h}_t = \textrm{GRU}(\mathbf{c}_{t}, \mathbf{\hat{h}}_t)
\end{equation}

The training loss of the merger is defined as follows:
\begin{align}	
	{\mathcal{L}_{\text{m}}} = \sum\nolimits_t{\sum\nolimits_i{\frac{{L({\hat m}_{ti}, y_{ti})}}{C{\left\| {{\mathbf{y}_t}} \right\|}_1}}}
	\label{cell-merge-loss}
\end{align}
where function $L$ has been defined in Eq.~\ref{BCE}, $C$ is the length of a predicted sequence and $y_{ti}$ denotes the ground-truth of cell's grid elements. $y_{ti}$ is $1$ if the $i^{th}$ grid element belong to the cell of time step $t$, otherwise $0$.

\subsection{Post-Processing}
	Through the merger, we can obtain the spanning of each table cell along the rows and columns. By combining the spanning information and the prediction results of the splitter, which contains the table row/column lines information, the bounding box coordinates of each table cell can be obtained. We match the text content with position to the table cells according to the IOU. The output for every table image finally contains coordinates of predicted cell bounding boxes along with cell spanning information and its content.

\section{Experiment}
\subsection{Dataset}
We use the publicly available table structure datasets — SciTSR~\cite{GraphTSR}, SciTSR-COMP~\cite{GraphTSR} and PubTabNet~\cite{EDD} to evaluate the effectiveness of our model. Statistics of these datasets are listed in Table~\ref{datasets}.

\begin{table}[!htp]
	\centering
	\renewcommand\arraystretch{1.2}
	\setlength{\tabcolsep}{6mm}
	
	\caption{Statistics of the datasets used for our experiments.}
	\label{datasets}
	\begin{tabular}{c|ccc}
		\hline
		Dataset 	& SciTSR & SciTSR-COMP & PubTabNet \\ \hline
		Train   	& 12k    & -   & 500k \\
		Validation  & -      & -   & 9k   \\
		Test    	& 3k     & 716 & 9k   \\ \hline
	\end{tabular}
\end{table}

1) \textbf{SciTSR}~\cite{GraphTSR} is a large-scale table structure recognition dataset, which contains 15,000 tables in PDF format as well as their corresponding high quality structure labels obtained from LaTeX source files. SciTSR splits $12,000$ for training and $3,000$ for testing. Furthermore, to reflect the model’s ability of recognizing complex tables, \cite{GraphTSR} extracts all the 716 complex tables from the test set as a test subset, called SciTSR-COMP. It's worth noting that SciTSR provides the text contents with positions for each table image, but not with being aligned with the table cells. However, in our model, we need the text position in each table cell to generate the labels of splitter. Therefore, we apply the data preprocessing~\footnote{\label{dataprocess}\url{https://github.com/ZZR8066/SciTSR}} to align the text information with the table cells.

2) \textbf{PubTabNet}~\cite{EDD} contains over 500k training samples and 9k validation samples. PubTabNet~\cite{EDD} annotates each table image with information about both the structure of table and the text content with position of each non-empty table cell. Moreover, nearly half of them are complex tables which have spanning cells in PubTabNet.

\subsection{Label Generation}
\begin{figure*}[htb]
	\centerline{\includegraphics[width=1.\linewidth]{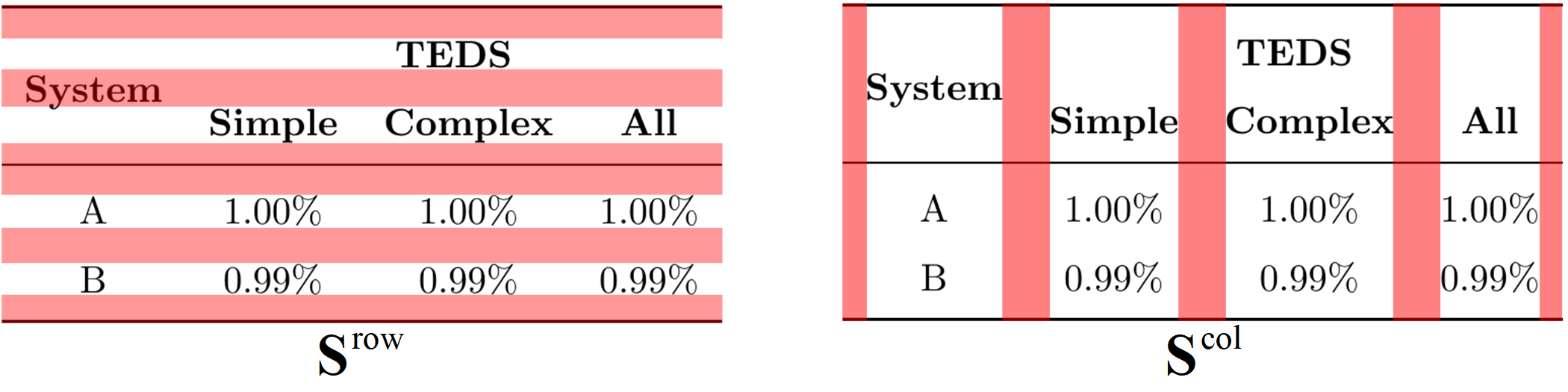}}
	\caption{Example of the ground-truth of table row/column separator map for the splitter. The red mask is the table row/column separator region.}
	\label{gt}
\end{figure*}

\textbf{Label of Splitter} We use the annotation, namely the text content with position being aligned to each table cell, to generate the ground-truth of the table row/column separator map $\mathbf{S}^{\text{row}}$/$\mathbf{S}^{\text{col}}$ for the splitter. The $\mathbf{S}^{\text{row}}$/$\mathbf{S}^{\text{col}}$ is designed to maximize the size of the separator regions without intersecting any non-spanning cell content, as shown in Figure~\ref{gt}. Different from traditional notion of cell separators, which for many tables are thin lines with only a few pixels. Predicting small regions is more difficult than predicting large regions. In the case of unlined tables, the exact location of the cell separator is ill-defined. 

\textbf{Label of Merger} Since we obtain the label of the splitter, we could divide the table into a set of basic table grids as shown in the upper-right of Figure~\ref{overview}. The original table structure annotation can provide which grids each table cell occupies in the basic table grid pattern. We arrange the table cells in a top-to-bottom and left-to-right way and use the grids occupied by each cell as the prediction target for a certain time step of the merger.

\subsection{Metric}
We use both F1-Measure~\cite{F1-Measure} and Tree-Edit-Distance-based Similarity (TEDS) metric~\cite{EDD}, which are commonly adopted in table structure recognition literature and competitions, to evaluate the performance of our model for recognition of the table structure.

In order to use the F1-Measure, the adjacency relationships among the table cells need to be detected. F1-Measure measures the percentage of correctly detected pairs of adjacent cells, where both cells are segmented correctly and identified as neighbors.

The TEDS metric was recently proposed in~\cite{EDD}. While using the TEDS metric, we need to present tables as a tree structure in the HTML format. Finally, TEDS between two trees is computed as:
\begin{equation}
	\text{TEDS}(T_a, T_b) = 1 - \frac{\text{EditDist}(T_a, T_b)}{\max (\lvert T_a \rvert, \lvert T_b \rvert)}
\end{equation}
where $T_a$ and $T_b$ are the tree structure of tables in the HTML formats. EditDist represents the tree-edit distance~\cite{TED}, and $\lvert T \rvert$ is the number of nodes in $T$.

\subsection{Implementation Details}
The modified ResNet-34~\cite{ResNet} as our backbone is pre-trained on ImageNet~\cite{ImageNet}. The number of FPN channels is set to $D = 256$. The pool size $R\times R$ of RoIAlign in vision module is set to $3\times 3$. The recognizer~\cite{WAP} is pre-trained on 35M table cell images, which are cropped from 500k table images in the PubTabNet dataset~\cite{EDD}, and the success rate of word predictions per table cell reaches 94.1\%. The BERT~\cite{Bert} we used is from the transformers package~\footnote{\label{tranformer}\url{https://github.com/huggingface/transformers}}. The hidden state dimension in the merger is set to 256.

The training objective of our model is to minimize the segmentation loss (Eq.~\ref{seg-row-loss}, Eq.~\ref{seg-col-loss}) and the cell merging loss (Eq.~\ref{cell-merge-loss}). The objective function for optimization is shown as follows:
\begin{equation}
	{O} = {\lambda _1}\mathcal{L}_s^{{\text{row}}} + {\lambda _2}\mathcal{L}_s^{\text{col}} + {\lambda _3}{\mathcal{L}_m}
\end{equation}
In our experiments, we set $\lambda _1 = \lambda _2 = \lambda _3 = 1$. We employ the ADADELTA algorithm~\cite{ADADELTA} for optimization, with the following hyper parameters: ${\beta _1} = 0.9$, ${\beta _2} = 0.999$ and $\varepsilon  = {10^{ - 9}}$. We set the learning rate using the cosine annealing schedule~\cite{sgdr} as follows:
\begin{equation}
	{\eta}_t = {\eta _{min}} + \frac{1}{2}({\eta _{max}} - {\eta _{min}})(1 + \cos (\frac{{{T_{cur}}}}{{{T_{max }}}}\pi ))
\end{equation}
where ${\eta}_t$ is the updated learning rate. ${\eta _{min}}$ and ${\eta _{max}}$ are the minimum learning rate and the initial learning rate, respectively. $T_{cur}$ and $T_{max}$ are the current number of iterations and the maximum number of iterations, respectively. Here we set ${\eta _{min}} = 10^{ - 6}$ and ${\eta _{max}} = 10^{ - 4}$.

Our model SEM is trained and evaluated with table images in original size. We use the NVIDIA TESLA V100 GPU with 32GB RAM memory for our experiments and the batch-size of 8. The whole framework was implemented using PyTorch.

\subsection{Visualization}
In this section, we visualize the segmentation results of the spliter and show how the merger recovers the table cells from the table grid elements through attention visualization.

\begin{figure*}[!htb]
	\centerline{\includegraphics[width=1.\linewidth]{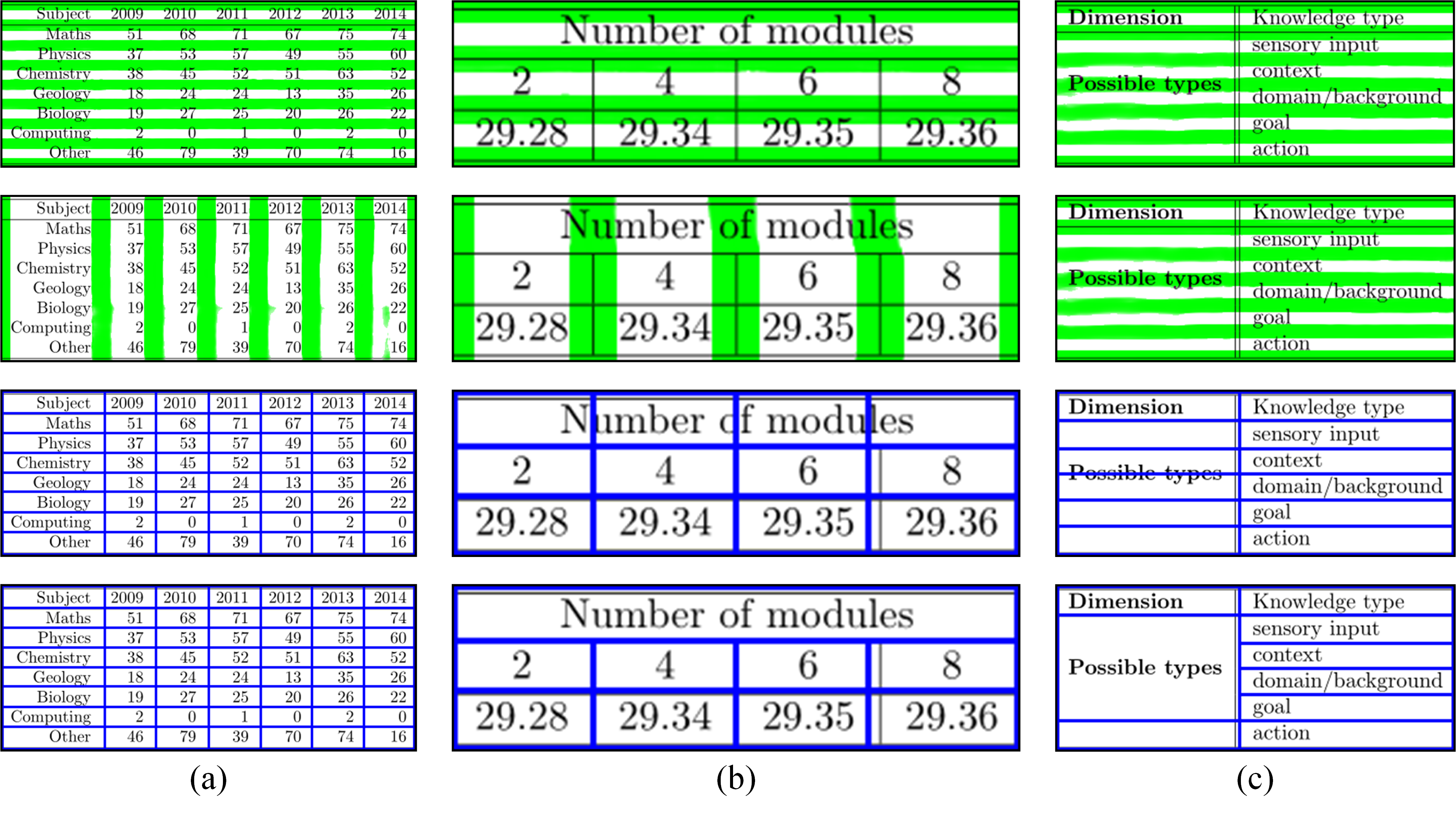}}
	\caption{The visualization results from our system on table images of the SciTSR dataset. \textbf{First Row:} the green masks are the segmentation results of the row segmenter in the splitter. \textbf{Second Row:} the green masks are the segmentation results of the column segmenter in the splitter. \textbf{Third Row:} the blue lines indicate the boundaries of the basic table grids which are extracted through post-processing from both row and column segmentation results. \textbf{Fourth Row:} the blue lines indicate the boundaries of the table cells which are the merged results from the merger.}
	\label{visualization_segmentation}
\end{figure*}

\textbf{Visualization of Splitter} We refer the potential regions of the table row (column) separators as the foreground as shown in Figure~\ref{gt}, and design the splitter which is actually a fully convolutional network (FCN) to predict the foreground in table images. As shown in the first two rows of Figure~\ref{visualization_segmentation}, we can obtain accurate segmentation results through the splitter. The fine grid structure of the table can be obtained by post-processing as shown in the third row of Figure~\ref{visualization_segmentation}. It is worth noting that the example table in Figure~\ref{visualization_segmentation} (a) is the simple table, while others are complex tables. We can find that the structure of the simple table has been recovered correctly through the splitter from the third row of Figure~\ref{visualization_segmentation}. However, the structure of complex tables is not complete and still needs to be processed. Therefore, we design the following embedder and merger to recover the structure of complex tables based on the outputs of the splitter.

\begin{figure*}[!htb]
	\centerline{\includegraphics[width=1.\linewidth]{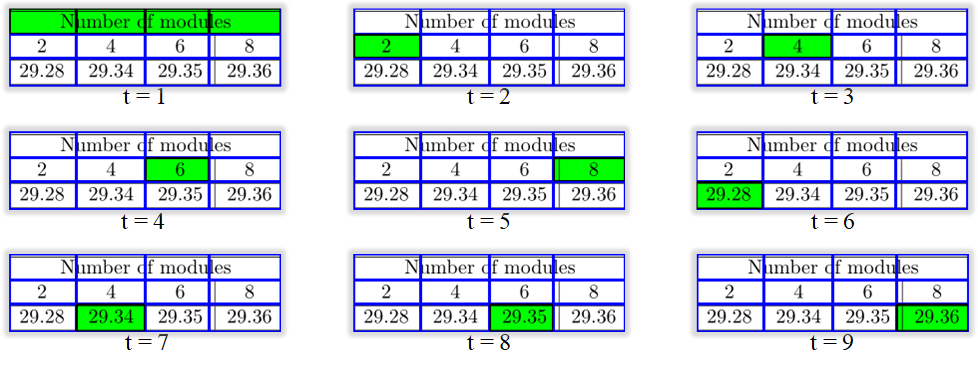}}
	\caption{The visualization of the attention mechanism in the merger on the table images of the SciTSR dataset. The blue lines are the prediction of table grid structure from the splitter. The green mask in the table image denotes which grid elements should be merged for each time step.}
	\label{visualization_merger}
\end{figure*}

\textbf{Visualization of Merger} In order to recover the table cells, we build the attention mechanism into our merger to predict which grid elements should be merged step by step. The merged result in each step is a binary map, and the table cell can be recovered by merging the elements that are 1 in the binary map. Taking the table of Figure~\ref{visualization_segmentation} (b) as a example, the attention mechanism is visualized in Figure~\ref{visualization_merger}. The cell with the content of “Number of modules” in Figure~\ref{visualization_merger} occupies the first row of basic table grids. Our merger correctly predicts the structure of this cell through the attention mechanism as shown in the first time step of Figure~\ref{visualization_merger}.

\subsection{Ablation Study}
In order to investigate the effect of each component, we conduct ablation experiments through several designed systems as shown in Table~\ref{ablation-systems}. The model is not modified except the component being tested.

\begin{table}[!htp]
	\centering
	\renewcommand\arraystretch{1.1}
	\setlength{\tabcolsep}{4mm}
	
	\caption{Comparison among systems from T1 to T4. Attributes for comparison include: 1) employing the splitter; 2) using the vision module (VM) in the  embedder; 3) using the text module (TM) in the embedder; 4) employing the merger.}
	\label{ablation-systems}
	\begin{tabular}{ccccc}
		\hline
		\multirow{2}{*}{System} & \multirow{2}{*}{Splitter} & \multicolumn{2}{c}{Embedder} & \multirow{2}{*}{Merger} \\
		&            & VM            & TM           &                         \\ \hline
		T1           & \checkmark          & -             & -            & -                \\
		T2           & \checkmark          & -             & \checkmark   & \checkmark       \\
		T3           & \checkmark          & \checkmark    & -            & \checkmark       \\
		T4           & \checkmark          & \checkmark    & \checkmark   & \checkmark       \\ \hline
	\end{tabular}
\end{table}

\begin{figure*}[!htb]
	\centerline{\includegraphics[width=.6\linewidth]{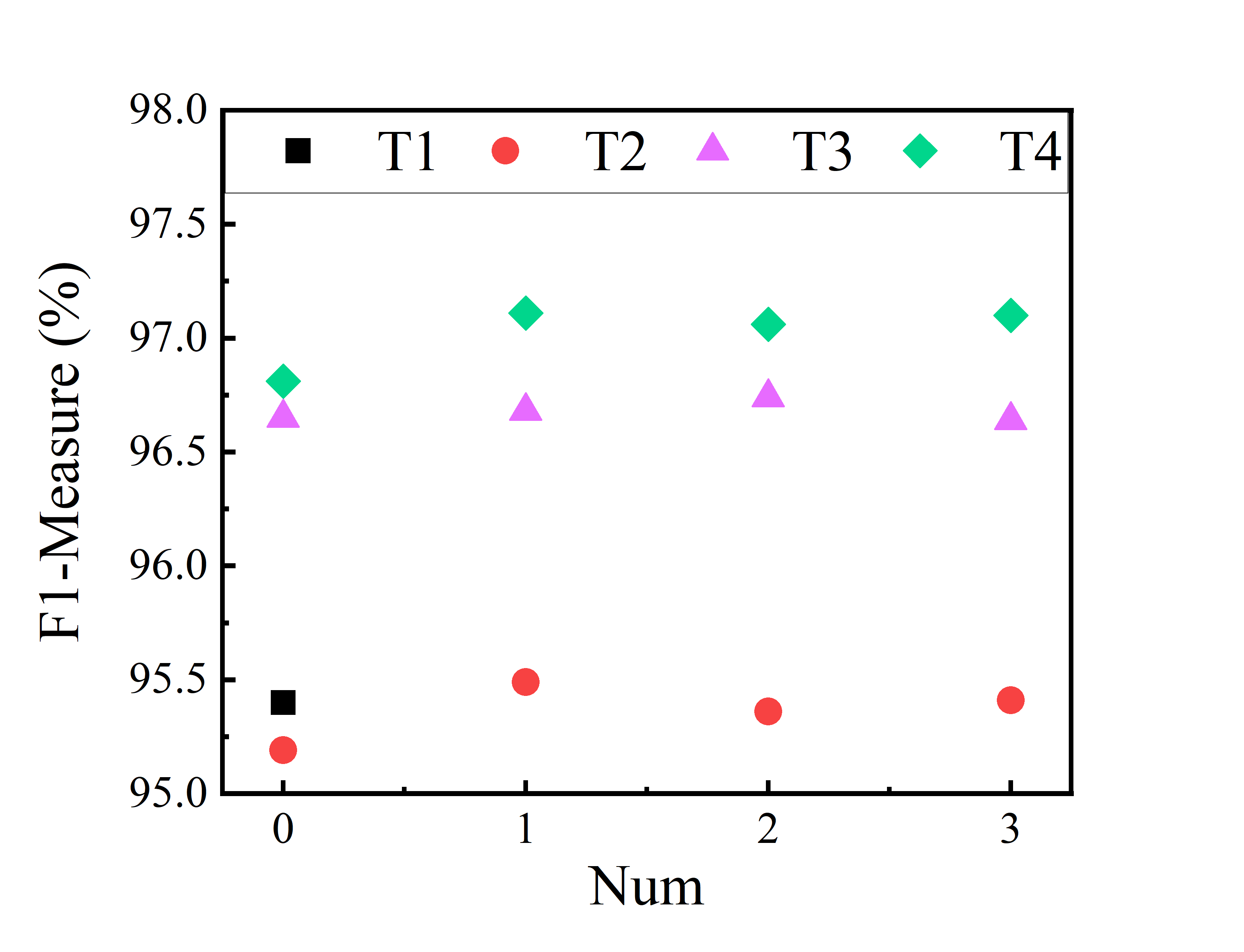}}
	\caption{Performance by varying number of transformer layers in T2, T3, T4 on the SciTSR test dataset.}
	\label{ablation-study-1}
\end{figure*}

\textbf{The Number of Transformer Layers} We measure the performance of $\text{T1-T4}$ with different numbers of transformer layers in the embedder. We try from 0 to 3 as shown in Figure~\ref{ablation-study-1}. When $\text{Num} = 0$ in Figure~\ref{ablation-study-1}, it means the transformer layer is removed. In the T3 configuration, only the vision module (VM) in the embedder is used to extract the visiual features to represent each grid element. Also there is not much gap whether the transformer layer is added or not. Through a series of convolutional layers, the backbone features P2 already has a certain receptive field. Therefore, it is not significant to add the transformer layers while the VM has pooled each grid features from P2. It is worth noting that when the Num is greater than 0, the performance of the designed system T2 outperforms the model without the transformer layer. This is because the transformer layer here can capture the semantical dependencies among all table grid elements. As our final system, T4 achieves the best result when $\text{Num} = 1$, so we set $\text{Num} = 1$ in subsequent experiments by default.

\begin{table}[!htp]
	\centering
	\renewcommand\arraystretch{1.1}
	\setlength{\tabcolsep}{3.3mm}
	
	\caption{Comparison of F1-Measure among different systems in Table~\ref{ablation-systems} on SciTSR and SciTSR-COMP datasets.}
	\label{ablation-studies}
	\begin{tabular}{c|c|c|c|c|c|c|c}
		\hline
		\multirow{2}{*}{System} & \multicolumn{3}{c|}{SciTSR} & \multicolumn{3}{c|}{SciTSR-COMP} & \multirow{2}{*}{FPS} \\ \cline{2-7} 
		& P       & R       & F1      & P       & R        & F1       & \\ \hline
		T1        & 96.69   & 94.15   & 95.40   & 93.81    & 96.06    & 89.77     & 16.47   \\
		T2        & 96.63   & 94.36   & 95.48   & 94.15    & 88.04    & 90.99     & 2.00	\\
		T3        & 97.40   & 95.97   & 96.68   & 96.52    & 93.82    & 95.15     & 3.65	\\
		T4        & 97.70   & 96.52   & 97.11   & 96.80    & 94.67    & 95.72     & 1.94	\\ \hline
	\end{tabular}
\end{table}

\textbf{The Effectiveness of the Merger} In Table~\ref{ablation-studies}, we show the F1-Measure of systems $\text{T1-T4}$ on SciTSR and SciTSR-COMP datasets. Almost $76.3\%$ of the tables are simple tables in SciTSR test dataset, and all are complex tables in the SciTSR-COMP dataset. The performance gap between T1 and other systems (T2-T4) is remarkable on the SciTSR dataset, but the gain is more significant on the SciTSR-COMP dataset, e.g., almost $6\%$ in F1-Measure from T1 to T4. This is because all table cells have only one table grid in the simple table, which means that the table grid structure is the table structure. However there are some table cells have more than one table grids in the complex table. Therefore, the designed system T1 can only process simple tables well by using splitter to predict the fine grid structure of table, and $\text{T2-T4}$ have the ability to recover the structure of the complex tables through the merger. The comparison of T1 with T2, T3, T4 on the SciTSR-COMP dataset demonstrates the effectiveness of the merger.

\begin{figure*}[!htb]
	\centerline{\includegraphics[width=1.\linewidth]{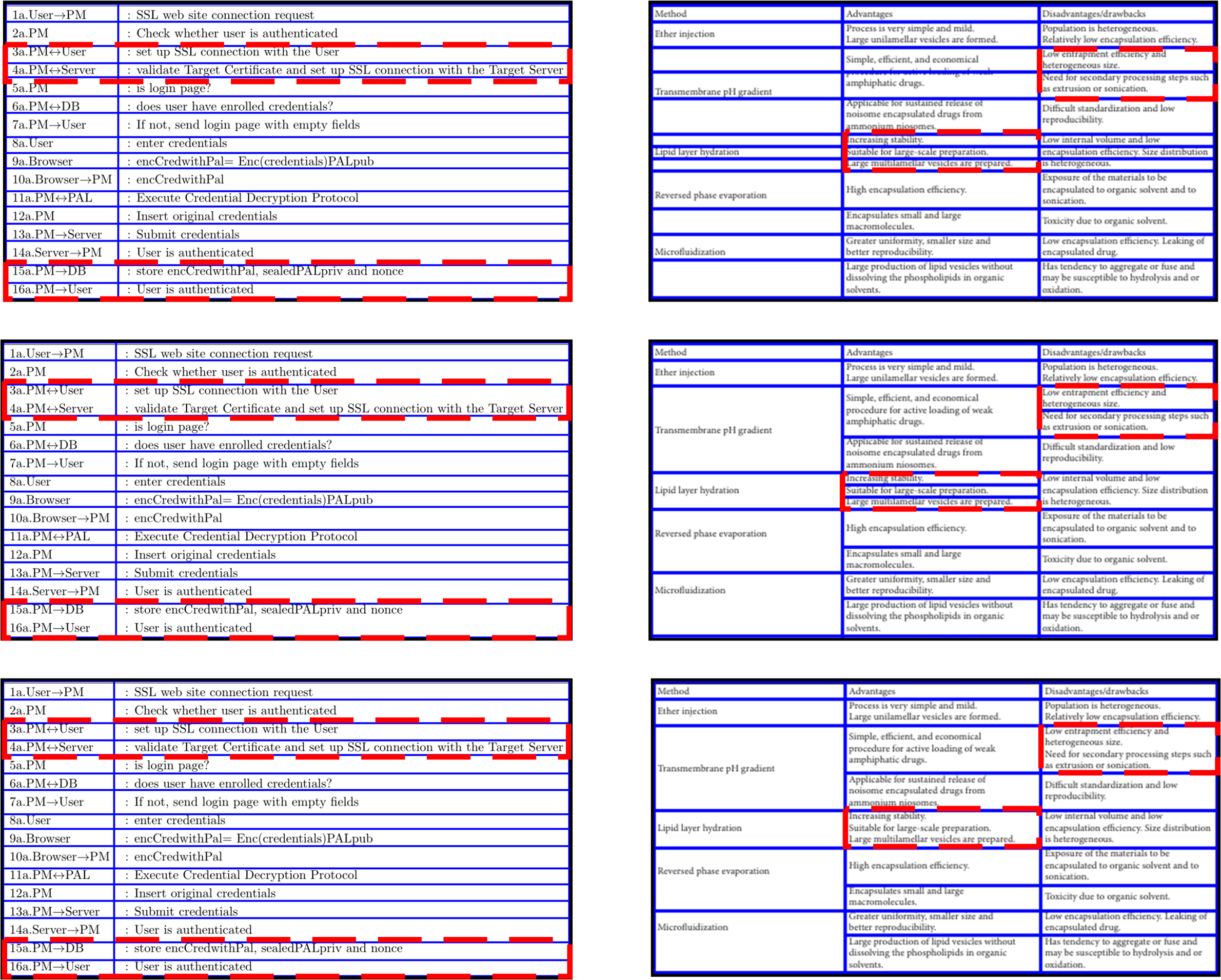}}
	\caption{The comparison results between designed system T3 and T4. The fisrt column is the results on the SciTSR dataset. The second column is the results on the PubTabNet dataset.  \textbf{First Row:} the predictions of the table grid structure from the splitter. \textbf{Second Row:} the predictions of the table structure from the T3 which only uses the vision module in the embedder. \textbf{Third Row:} the predictions of the table structure from the T4 which uses both the vision module and text module in the embedder. Note that the predictions of table grid structure in systems T3 and T4 are the same, and the predictions of table structure in the third row are all totally correct. The red dash boxes denote the different predictions between T3 and T4.}
	\label{visualization_comparison}
\end{figure*}

\textbf{Vision and Language Modalities} In order to evaluate the effect of each modality, we design the systems T2, T3, T4 as shown in Table~\ref{ablation-systems}.
Each system uses vision module (VM), text module (TM) or both in the embedder. The experiment results on SciTSR and SciTSR-COMP are shown in Table~\ref{ablation-studies}. Compared with T4, systems T2 and T3 that only use TM or VM are sub-optimal. When both TM and VM are used, the system (T4) performance reaches the best. As shown in Figure~\ref{visualization_comparison}, although the predictions of table grid structure from the splitter in both T3 and T4 are the same, the T3 system which only uses VM is more unstable comparing with T4 which uses both VM and TM in the embedder.

\textbf{The Efficiency of Each Component} In order to investigate the efficiency of each component, we compare the frames per second (FPS) of T1-T4 systems as shown in Table~\ref{ablation-studies}.  From T1 to T2-T4 systems, the speed of the systems is much slower. This is because as the number of table cells increases, the decoding steps of the merger increases. The reason why T2 and T4 are slower than T3 is that the former uses a recognizer to recognize the content in the basic table grid and applies the BERT to extract the corresponding textual features.

\begin{table}[!htb]
	\centering
	\renewcommand\arraystretch{1.1}
	\setlength{\tabcolsep}{2.2mm}
	
	\caption{A performance comparison between our method and other state-of-the-art methods on the SciTSR and SciTSR-COMP datasets.}
	\label{scitsr-comparison}
	\begin{tabular}{c|c|c|c|c|c|c|c}
		\hline
		\multirow{2}{*}{Method} & \multicolumn{3}{c|}{SciTSR} & \multicolumn{3}{c|}{SciTSR-COMP} & \multirow{2}{*}{FPS} \\ \cline{2-7} 
										   & P       & R       & F1      & P      & R        & F1 		& 		\\ \hline
		Adobe~\cite{DeepDeSRT}	           & 82.9    & 79.6    & 81.2    & 79.6   & 73.7     & 76.5  	& -		\\
		TabbyPDF~\cite{TabbyPDF}           & 91.4    & 91.0    & 91.2    & 86.9   & 84.1     & 85.5 	& -		\\
		DeepDeSRT~\cite{DeepDeSRT}         & 89.8    & 89.7    & 89.7    & 81.1   & 81.3     & 81.2 	& 20.88				\\
		GraphTSR~\cite{GraphTSR}           & 93.6    & 93.1    & 93.4    & 94.3   & 92.5     & 93.4 	& - 									\\
		TabStruct-Net~\cite{TabStructNet}  & 92.7    & 91.3    & 92.0    & 90.9   & 88.2     & 89.5 	& 0.77				\\ \hline
		${\text{T1}^{\,\,\,}}$		       & 96.69   & 94.15   & 95.40   & 93.81  & 96.06    & 89.77 	& 16.47				\\
		${\text{SEM}^{\,\,\,}}$    		   & 97.70   & 96.52   & \textbf{97.11}   & 96.80    & 94.67  & \textbf{95.72} & 1.94 \\ \hline
	\end{tabular}
\end{table}

\subsection{Comparison with State-of-the-art Methods}
\begin{figure*}[!htb]
	\centerline{\includegraphics[width=1.\linewidth]{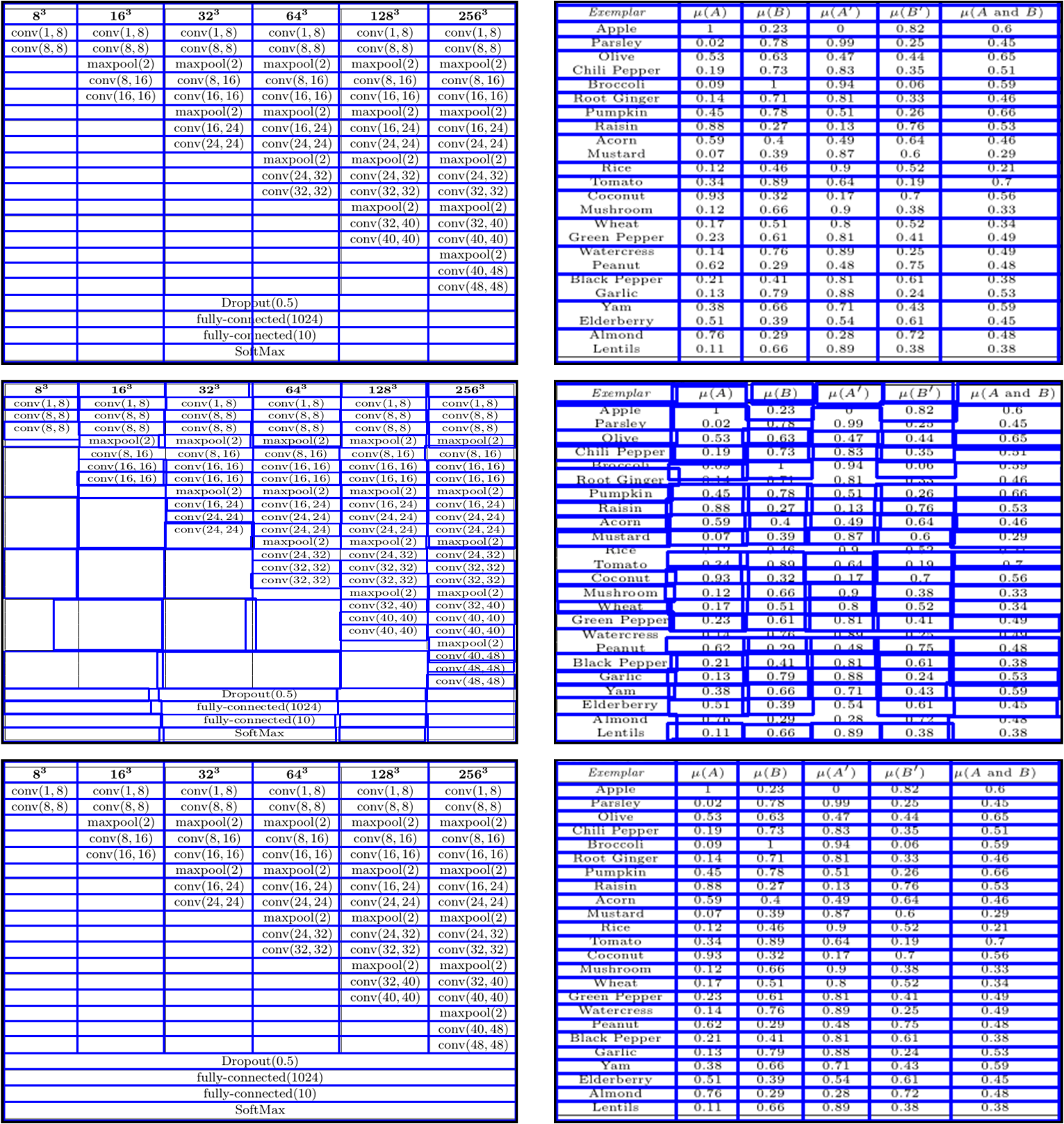}}
	\caption{Some structure recognition results of our and other methods on table images of the SciTSR dataset. The blue lines denote the prediction of table structure. \textbf{First Row:} the results of the DeepDeSRT. \textbf{Second Row:} the intermediate cell detection results of the TabStruct-Net. \textbf{Third Row:} the results of our method. The predictions of table structure in the third row are all correct.}
	\label{visualization_SciTSR}
\end{figure*}

We compare our method with other state-of-the-art methods on both SciTSR and SciTSR-COMP datasets. The results are shown in Table~\ref{scitsr-comparison}. Our model is trained and tested with default configuration. Comparing with other methods~\cite{GraphTSR,DeepDeSRT,TabStructNet}, our method achieves state-of-the-art. Similar to DeepDeSRT~\cite{DeepDeSRT}, T1 is actually a segmentation model. We take full consideration of the extremely unbalanced number of foreground and background pixels in the segmentation masks and design a more reasonable segmentation loss to penalize the model during training in Eq.~\ref{seg-row-loss}~\ref{seg-col-loss}, which makes the performance of T1 significantly better than DeepDeSRT. It is worth noting that GraphTSR~\cite{GraphTSR} needs the text position in table cells during both the training and testing stage, while our method only takes table images as input during inference. Although the comparison between GraphTSR and our method is not fair, our method still outperforms it to a certain extend. The TabStruct-Net~\cite{TabStructNet} applies a detection network~\cite{MaskRCNN} to detect individual cells in a table image, however, it fails to capture empty cells accurately due to the absence of cell content. Our SEM obtains the bounding boxes of cells based on table lines detection, which makes the prediction of empty cells less difficult. In addition, the embedder makes full use of the visual and textual modalities, and the merger enables the model to process complex tables, which ultimately makes our method achieve state-of-the-art. Some table structure recognition results of our and other methods are shown in Figure~\ref{visualization_SciTSR}.

\subsection{ICDAR 2021 Competition on Scientific Literature Parsing, Task-B}
ICDAR 2021 Competition on Scientific Literature Parsing, Task-B~\footnote{\label{pubtabnet-competition}\url{https://aieval.draco.res.ibm.com/challenge/40/overview}} is organized by the IBM company in conjunction with IEEE ICDAR 2021. This competition aims to drive the advances in table recognition. Different from the table structure recognition task, we need to recognize not only the structure of the table, but also the content within each cell. Through our method, we can not only predict the structure of the table, but also obtain the position of each cell. Inspired by~\cite{ABCNet,MaskTextSpotterV1,MaskTextSpotterV2,MaskTextSpotterV3}, we use the RoIAlign to pool the features of table cells and append an attention-based recognizer~\cite{WAP} to recognize the content in table cells. Note that the modified models are trained in an end-to-end manner. The single model results of our methods are shown in Table~\ref{e2e-comparsion}.

\begin{table}[!htp]
	\centering
	\renewcommand\arraystretch{1.1}
	\setlength{\tabcolsep}{4mm}
	
	\caption{The performance of table recognition on PubTabNet validation set.}
	\label{e2e-comparsion}
	\begin{tabular}{c|c|c|c|c}
		\hline
		\multirow{2}{*}{Method} & \multicolumn{3}{c|}{TEDS} & \multirow{2}{*}{FPS}  \\ \cline{2-4} 
		& Simple  & Complex  & All &         \\ \hline
		${\text{T3} + \text{Recognizer}}$   & 94.7 & 92.1 & 93.4 & 1.81 \\
		${\text{T4} + \text{Recognizer}}$	& 94.8 & 92.5 & 93.7 & 1.23 \\ \hline
	\end{tabular}
\end{table}

Based on the configuration of T3 with a recognizer, we divide our model into three sub-networks, splitter, merger and newly added recognizer, adopting multi-model fusion for each sub-network. Finally, we combine the training set with the validation set for training. The results of the competition are shown in Table~\ref{pubtabnet-competition}. Our team is named USTC-NELSLIP, and we won the first place of complex tables and third place of all tables.

\begin{table}[!htp]
	\centering
	\renewcommand\arraystretch{.9}
	\setlength{\tabcolsep}{4mm}
	
	\caption{Table recognition competition results on PubTabNet final evaluation data set.}
	\label{pubtabnet-competition}
	\begin{tabular}{|c|c|c|c|}
		\hline
		\multirow{2}{*}{Team Name} & \multicolumn{3}{c|}{TEDS}                        \\ \cline{2-4} 
		& Simple         & Complex        & All            \\ \hline
		Davar-Lab-OCR              & 97.88          & 94.78          & \textbf{96.36} \\ \hline
		VCGroup                    & \textbf{97.90} & 94.68          & 96.32          \\ \hline
		\textbf{USTC-NELSLIP}      & 97.60          & \textbf{94.89} & 96.27          \\ \hline
		YG                         & 97.38          & 94.79          & 96.11          \\ \hline
		DBJ                        & 97.39          & 93.87          & 95.66          \\ \hline
		TAL                        & 97.30          & 93.93          & 95.65          \\ \hline
		PaodingAI                  & 97.35          & 93.79          & 95.61          \\ \hline
		anyone                     & 96.95          & 93.43          & 95.23          \\ \hline
		LTIAYN                     & 97.18          & 92.40          & 94.84          \\ \hline
		EDD						   & 91.20  & 85.40  & 88.30 \\ \hline
	\end{tabular}
\end{table}

\begin{figure*}[!htb]
	\centerline{\includegraphics[width=1.\linewidth]{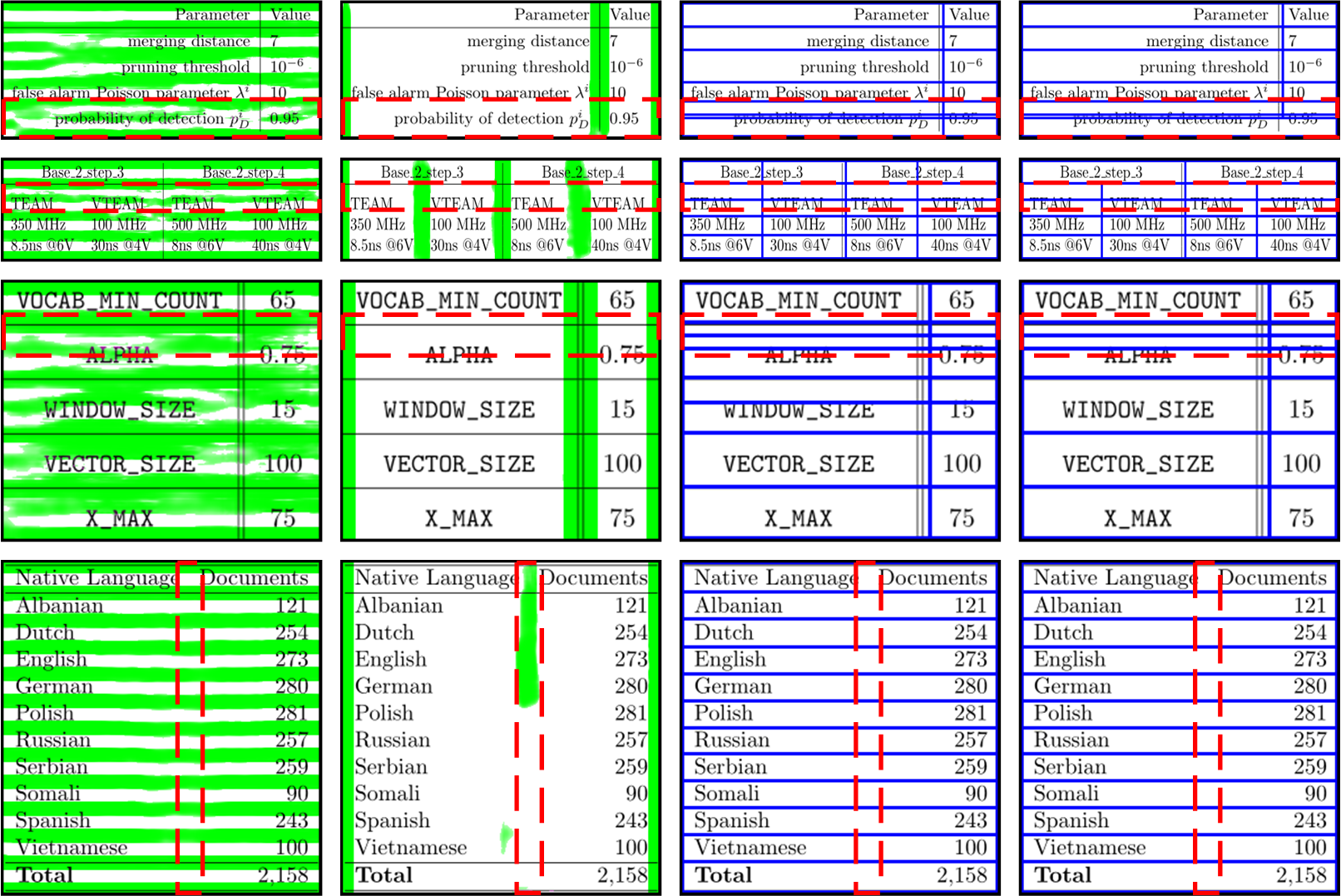}}
	\caption{Some incorrect table structure recognition results of our method. \textbf{First Column}: the row segmentation results of the splitter. \textbf{Second Column}: the column segmentation results of the splitter. \textbf{Third Column}: the predictions of the table grid structure from the splitter. \textbf{Fourth Column}: the final prediction results of our method. The red dash boxes denote the incorrect prediction results.}
	\label{error_analysis}
\end{figure*}

\subsection{Error Analysis}
In this section, we show some incorrect table structure recognition results of the SEM as shown in Figure~\ref{error_analysis}. Our splitter occasionally misses or overcuts the basic table grids when the blank space between cells is too large. In the training phase, the merger predicts the spanning information of cells on the correct basic table grid pattern. Therefore, in the inference phase, once the splitter predicts incorrect results, it is difficult for the merger to fix them.

\section{Conclusion}
In this study, we proposed a new method for the table structure recognition, SEM. The proposed method takes images as input with no dependency on meta-information or OCR. It mainly contains three components including splitter, embedder and merger. We first split table images into a set of basic table grids. Then the embedder is used to extract the feature representations of each grid element. Finally, we use the merger with the attention mechanism to predict which grid elements should be merged to recover the table cells. The final table structure can be obtained by parsing all table cells. The method can not only process simple tables well, but also the complex tables. We demonstrate through visualization and experiment results that the attention mechanism built in the merger performs well in predicting which grid elements belong to each cell. To our best knowledge, this is the first time to take a full consideration of the textual information in table images and design the embedder to extract both the visual and the textual features. The ablation studies prove the effectiveness of our embedder. Our method achieves state-of-the-art on both SciTSR and SciTSR-COMP datasets. Based on our method, we won the first place of complex tables and third place of all tables in Task-B of ICDAR 2021 Competition on Scientific Literature Parsing.

\bibliography{reference}

\end{document}